\documentclass[10pt]{IEEEtran}
\usepackage[T1]{fontenc}
\usepackage[latin9]{inputenc}
\usepackage{units}
\usepackage{amsmath,lipsum}
\newcommand\numberthis{\addtocounter{equation}{1}\tag{\theequation}}
\usepackage{stmaryrd}
\usepackage{stackrel}
\usepackage{graphicx}
\usepackage{wasysym}
\usepackage{amssymb}
\usepackage{lineno}
\usepackage{multicol}
\usepackage{geometry}
\usepackage{cleveref} 
\ifCLASSINFOpdf
\else
\fi

\begin{document}
%
\title{Frequency Aware Face Hallucination Generative Adversarial Network with Semantic Structural Constraint}
%
%
%

\author{Shailza~Sharma,
        Abhinav~Dhall,
        and~Vinay~Kumar
\thanks{S. Sharma and V. Kumar are with the Department
of Electronics and Communication Engineering, Thapar Institute of Engineering and Technology, India,
e-mail: (ssharma\_phd18@thapar.edu; vinay.kumar@thapar.edu).}
\thanks{A. Dhall is with Department of Human Centred Computing in Monash University, Australia,
e-mail: (abhinav.dhall@monash.edu)}}
\maketitle

\begin{abstract}
In this paper, we address the issue of face hallucination.
Most current face hallucination methods rely on two-dimensional
facial priors to generate high resolution face images from low resolution
face images. These methods are only capable of assimilating global
information into the generated image. Still there exist some inherent
problems in these methods; such as, local features, subtle structural
details and missing depth information in final output image. Present
work proposes a Generative Adversarial Network (GAN) based novel progressive
Face Hallucination (FH) network to address these issues present among
current methods. The generator of the proposed model comprises of
FH network and two sub-networks, assisting FH network to
generate high resolution images. The first sub-network leverages on
explicitly adding high frequency components into the model. To
explicitly encode the high frequency components, an auto encoder is
proposed to generate high resolution coefficients of Discrete Cosine
Transform (DCT). To add three dimensional parametric information into
the network, second sub-network is proposed. This network uses
a shape model of 3D Morphable Models (3DMM) to add
structural constraint to the FH network. Extensive experimentation results in the paper shows that the proposed model outperforms the state-of-the-art
methods.
\end{abstract}


%

\section{Introduction}
To obtain the high resolution image from its equivalent low resolution counterpart is referred as Super Resolution (SR).  Particularly,  applying super resolution on faces is known as Face Hallucination (FH). High resolution facial images are
widely required in many image processing applications, such as facial
emotion detection \cite{han2018face}, pedestrian re-identification
\cite{nasrollahi2014super}, facial alignment \cite{bulat2017far},
face recognition \cite{lu2018robust} and face identification \cite{taigman2014deepface}.

Based on deep learning, researchers have effectively applied numerous
algorithms to solve face super resolution problem \cite{chen2018fsrnet}. However, the first drawback of majority of
these algorithms is that they rely on two dimensional facial priors to recover structural details of facial images, such as, Grm et al. \cite{grm2019face}
proposed a face Super Resolution (SR) algorithm, where the identity priors are incorporated
at multiple stages of Convolutional Neural Networks (CNN) to perform image hallucination. Progressive
facial attention loss is proposed by Kim et al.\cite{kim2019progressive}
to incorporate facial attributes in the generated SR face images.
Yu et al.\cite{yu2018face} used coarse SR image from the intermediate
stage to extract heatmaps by passing that image in an UNet architecture.
Extracted heatmaps are then merged with coarse SR image to generate
final SR image. All the above mentioned methods used 2D priors to
guide the deep learning models to generate the high resolution face
images. The information extracted from the 2D priors is only capable
of incorporating global features in an output image. Still the local
features or the subtle structural details like skin irregularities,
wrinkles and depth details are missing in the final output image. To embed the above mentioned features in the generated image, we have proposed a progressive Face Hallucination (FH) network. An auxiliary sub-network is employed in the proposed FH network by using 3D Morphable Models (3DMMs) \cite{kittler20163d} to embed structural information in the output image.
3DMMs are the three
dimensional meshes of face images used to reconstruct a 3D image from
its 2D counterpart using shape and texture models of Principal Component
Analysis (PCA). To add the semantic structural constraint to proposed
FH model, we are utilizing the PCA shape model of 3DMM \cite{huber2016multiresolution}.
The shape components of PCA shape model constitute different face
parameters, like, the first shape component signifies the shape of
face (slim, chubby or round etc). And the secondary shape components
accounts for the more finer details (wrinkles, face irregularities,
face depth etc) of face images. These three dimensional meshes with
face parameters are rendered as two dimensional points on the face
image. The obtained 3D points fitted on 2D images act as a target
images for our auxiliary network. This auxiliary network act
as a supervision network to add structural constraint to our face hallucination network.

Second drawback of Super resolution methods based on generative adversarial networks
\cite{ledig2017photo,wang2018esrgan} is that the quantitative results
produced by these methods have less values as compared to other deep
learning based methods due to missing frequency details. To overcome
this drawback, a sub-network comprised of an auto encoder is added
along with the proposed FH network to explicitly add high frequency facial
details from an face image. Discrete Cosine Transform (DCT) based
feature maps with high resolution are generated using this network.
Frequency domain loss is calculated between the generated DCT feature
maps and ground truth DCT feature maps which is used to train the  sub-network. This loss function will guide our FH network to produce
output images with high quantitative values. To embed high feature DCT
maps in the FH network, we used an Inverse Discrete Cosine Transform (IDCT) block. This IDCT block
converts the frequency domain feature maps into the spatial domain
feature maps. Then the converted feature maps are merged with the
output feature maps of FH network, supervising our network to
produce images with high frequency details.

Main contributions of presented work are as follows:
\begin{enumerate}
\item GAN based progressive face hallucination network is proposed. The
generator network in the proposed network comprises of FH network
and two sub-networks, assisting FH network to generate high
resolution face image. The FH network consists of Hierarchical Feature Extraction Module and Computationally Efficient Channel based Attention Module. The hierarchical feature extraction module helps the model to learn hierarchical as well as primitive information present in the image. Where as channel based attention block is used to add channel-wise attention in the network and reduce the computational complexity of the network.

\item First sub-network produces high resolution DCT feature maps. This
network supervises FH network to produce images with high frequency
details. Proposed frequency domain based loss, assists our
face hallucination network to reflect the quality of resultant images. IDCT block is used
at the end of this network to convert the frequency domain feature
maps to spatial domain and merge them in the FH network.
\item An auxiliary sub-network generates a high resolution
2D image with 3D parameters fitted on it. This network adds, structural constraint to our FH network, producing
face images with semantic facial details, like skin irregularities,
wrinkles and depth information etc.
\item Experiments on facial benchmark datasets reflect superior performance
over recent state-of-the-art methods.
\end{enumerate}

\section{Related Work}
We have discussed the framework and some recent works related to image super resolution, face hallucination and frequency domain based deep learning methods in this section. High resolution images are the most essential requirement in many
computer vision applications. From the last few years, many approaches,
including frequency-domain methods \cite{nasrollahi2014super}, Bayesian
methods \cite{sun2003image,shan2008fast}, interpolation methods \cite{lehmann1999survey},
regularization based methods \cite{farsiu2004fast,belekos2010maximum}
and edge statistics methods \cite{shan2008fast} have
been proposed for image super resolution problem. But in recent times,
convolutional neural networks and generative adversarial networks
have captivated extensive attention from researchers to perform the
image super resolution task. 

SRCNN, the first CNN based implementation was proposed by Dong et al.
\cite{dong2014learning}. Based on this pioneer work, more CNN based
architectures are explored for image super resolution \cite{dong2016accelerating,shi2016real}.
VDSR, a deep CNN architecture, based on residual learning is proposed
by Kim et al. \cite{kim2016accurate} to perform image super resolution.
Batch Normalization layers are removed from the residual networks
to increase the training accuracy of image super resolution system
by Lim et al. \cite{lim2017enhanced}. Deep recursive network structure
(DRCN) is proposed by Jiwon et al. \cite{kim2016deeply} to extract
feature maps repetitively from the same filter structure to reduce the
count of parameters. Sometimes recursive networks induce instability
in the networks due to vanishing gradient problem. Skip connections
are used by the author in the SR recursive network to overcome this
problem. For large upscaling factors, Lai et al.\cite{lai2017deep}
proposed a deep CNN, named LapSRN, where features are upscaled using gradual
upscaling technique. This network contains two branches, where first
branch generate feature maps and the second branch is used for reconstruction.
Residual blocks and dense blocks are combined by Zhang et al. \cite{zhang2018image} to extract
the hierarchical features from each convolution layer along with the channel attention layer. For better correlation
between the extracted features from the intermediate layers Dai et
al. \cite{dai2019second} proposed a network (SAN) based on second
order attention mechanism. This network consists of three modules:
1) Adaptive channel-wise feature learning module, 2) NLRG- for long
term memory, and 3) LSRAG- for precise feature representation.

High frequency feature extraction capability of networks is diminished
by using loss functions that depend on pixel-wise differences \cite{bruna2015super}.
Therefore, SRGAN \cite{ledig2017photo}, the first GAN based network,
addresses this problem by incorporating the loss function based on feature-wise
differences. SRGAN is modified by Wang et al. \cite{wang2018esrgan}
to generate more realistic images. The architecture integrates the
dense residual blocks in the generator architecture with relativistic
\cite{zhang2018super} discriminator. Feature based discriminator
is proposed by Park et al. \cite{park2018srfeat} which helps the
model to reduce noisy artifacts and produces SR images with
high structural features.

To generate a high resolution face image, Cyang et al. \cite{yang2013structured}
integrated the image gradients obtained from specific face components.
Edges, smooth regions and LR exemplar images, selected on the basis
of pose and landmark detection constitute the face components. CNN
based architecture, SICNN is proposed by Zhang et al. \cite{zhang2018super},
where identity enhanced high resolution face images are generated
using identity loss function. Generated SR image and HR image are
applied to the face recognition CNN network and the features extracted
from this network are used to calculate the identity loss function.
Parsing maps and facial landmarks are used with the high resolution
face images to train GAN based network for generating SR face images
by Chen et al. \cite{chen2018fsrnet}. 

Supervised discriminator with
two inputs (SR image and the extracted features of SR image obtained
from pre-trained face recognition system) is proposed by Zhang et
al. \cite{zhang2020supervised} for face super resolution. This face
super resolution system is capable of generating face images with
very fine textural details. Two CNN branches- one with the facial
structural information (alligned heatmaps of nose, eyes, skin and
chin) and the other for the face super resolution are aggregated by
Yu et al. \cite{yu2018face}. Kalarot et al. \cite{kalarot2020component}
used a segmentation network to obtain three facial features: hair,
skin and other parts of face. These facial feature heat maps are merged
with the input image and passed to two stage image super resolution
network to obtain SR images. Image SR network based on spatial attention
is proposed by Chen et al. \cite{chen2020learning}. This network
permit the CNN layers to learn more parameters from the high textural
regions using face attention units. SRGAN \cite{ledig2017photo} is
improved by Wang et al. \cite{wang2020improved} by replacing the
residual blocks with dense blocks for face super resolution. In addition,
spectral normalization is introduced in the network to improve its
training efficiency. 

Discrete Cosine Transform based methods represents the feature maps in the frequency domain rather than representing them in spatial domain. For various applications, CNNs are being trained in frequency domain, such as, Zhang et al. \cite{zhang2020deep} extended the idea of DCT coefficients
and presented median filtering forensics approach which
was based on a convolutional neural network (CNN) with
an adaptive filtering layer (AFL) built in the discrete cosine
transform (DCT) domain. Meanwhile, Verma et al. \cite{verma2018dct}
addressed the problem of classifying images based on the
number of JPEG compressions they have undergone, by utilizing
deep convolutional neural networks in DCT domain.
For the task of super resolution, Islam et al. \cite{islam2012single} used directional
fourier phase feature components to adaptively learn
the regression kernel based on local covariance to estimate
the high-resolution image. \cite{li2018frequency} presented a frequency domain
neural network where convolutions in the spatial domain
was cast as products in the frequency domain and nonlinearity
was cast as convolution in the frequency domain.
Guo et al. \cite{guo2019adaptive} integrated DCT into the network structure as
a Convolutional DCT (CDCT) layer and formed DCT Deep
SR (DCT-DSR) network.

Our approach presents a novel way to integrate spatial and frequency domain components and add structural constraint to the resultant image by using 3D parametric information. Detailed explanation of proposed methodology is present in next section.

\section{Methodology}

\begin{figure*}[ht]
\centering

\includegraphics[scale=0.65]{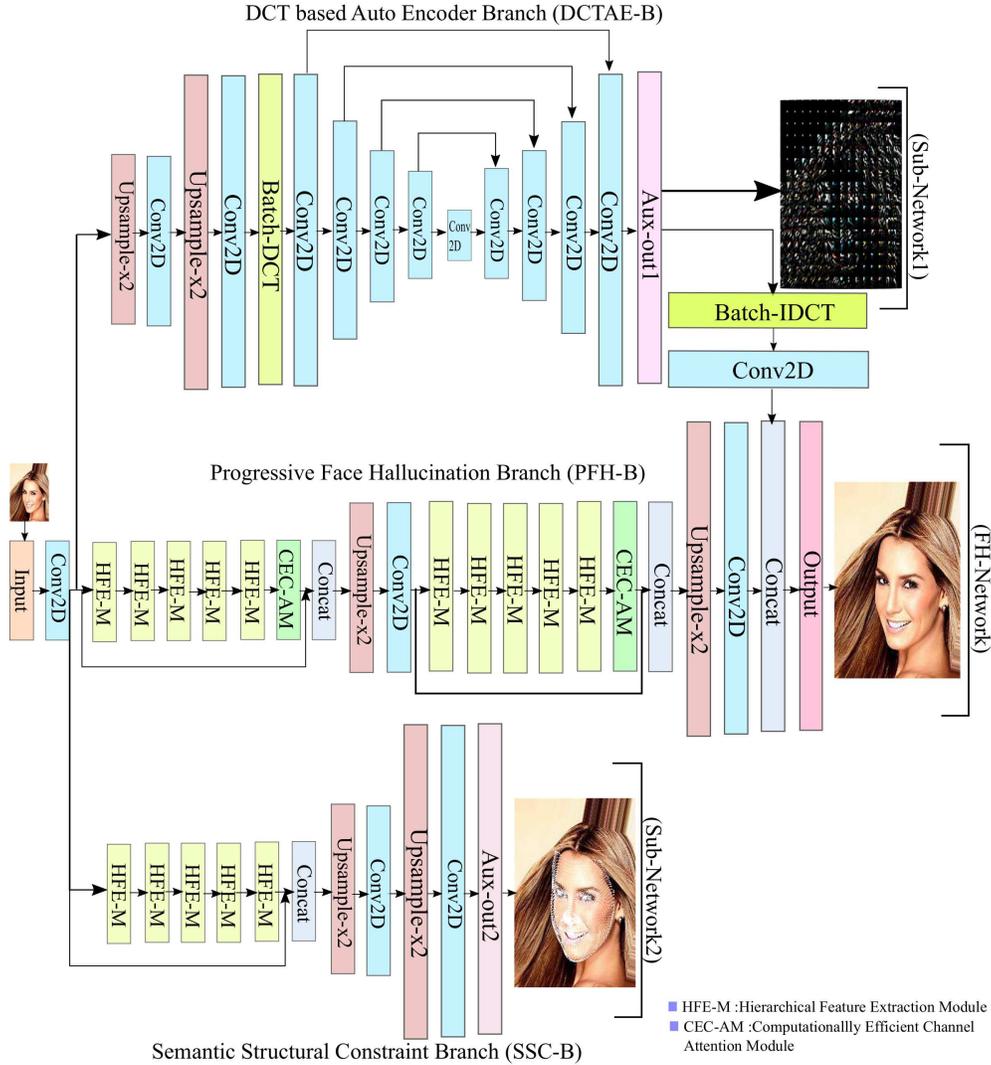}
\caption{Proposed Generator architecture with three branches: 1) FH network- progressive face hallucination branch from where resultant output image is generated, 2) Sub-network1- a DCT based encoder network to add high frequency components in the output image, and 3) Sub-network2- semantic structural constraint branch to add 3D parametric information in the generated image.     \label{fig:Generator-architecture}}
\end{figure*}

As depicted in the figure \ref{fig:Generator-architecture}, generator network of the proposed
architecture consist of three branches:

i) Progressive Face Hallucination
Branch (PFH-B), powerfully built with a combination of cascaded hierarchical
feature extraction module and channel based attention module, 

ii) Semantic Structural
Constraint Branch (SSC-B), serves as a supervision network by constraining
PFH-B to generate resultant images with three dimensional parametric
feature information, and 

iii) DCT based Auto Encoder Branch (DCTAE-B),
compelling PFH-B to produce images with high frequency details.

Basically, the objective of proposed face hallucination model is to find the mapping function $\mathcal{F}_{\theta_{fh}}$
(refer eq. \ref{eq:map}) to obtain a high resolution face image
($HR_{fi}$) from its low resolution counterpart ($LR_{fi}$).

\begin{equation}
\mathcal{F}_{\theta_{fh}}=LR_{fi}\shortrightarrow HR_{fi}\label{eq:map}
\end{equation}

where, $\theta_{fh}$ are the parameters learned throughout the mapping process. To minimize the distance between $HR_{fi}$ and its counter
$LR_{fi}$, proposed face hallucination network employs the combination of pixel based loss, feature based
loss, structured parametric loss and DCT based loss and updates the
learnable parameters during the training process. The output face images generated by face hallucination network is fed to discriminator network \cite{ledig2017photo} along with the high resolution face images. In the proposed architecture, discriminator is acting as a binary classifier, and trained in such a way that it classifies the ground-truth face images as label $1$ and generated face images as label $0$. On the contrary, generator is trained to trick discriminator by generating the face images similar to high resolution face images. Comprehensive training of both these networks will lead to the generation of high resolution face images.  

Detailed explanation of components employed in the generator network are discussed next.

\subsection{Progressive Face Hallucination Branch}

As illustrated in figure \ref{fig:Generator-architecture}, PFH-B
uses a progressive upscaling technique where at every stage the input
image is upscaled by the factor of $2$. Further, each stage
is divided into three phases. First phase is Elementary Characteristic
Extraction Phase (ECE-P), where a low resolution face image is passed
through a convolution layer to extract the elementary characteristics
of an image. The feature maps obtained from the first phase are applied
to the second phase which is Hierarchical Feature Extraction with
Channel Attention Mechanism (HEF-CA). This is the most crucial component
of PFH-B. HEF-CA is composed of two main blocks - Hierarchical Feature
Extraction Block and Computationally Efficient Channel based Attention
module, which are explained in detail in the following subsection.
Output feature maps obtained from the HEF-CA are applied to the reconstruction
phase, where the features maps are upsampled by the factor $\times2$
using subpixel convolution layer \cite{shi2016real} to get the final
output image. 

\subsubsection{Hierarchical Feature Extraction Module}

\begin{figure}
\centering

\includegraphics[scale=0.4]{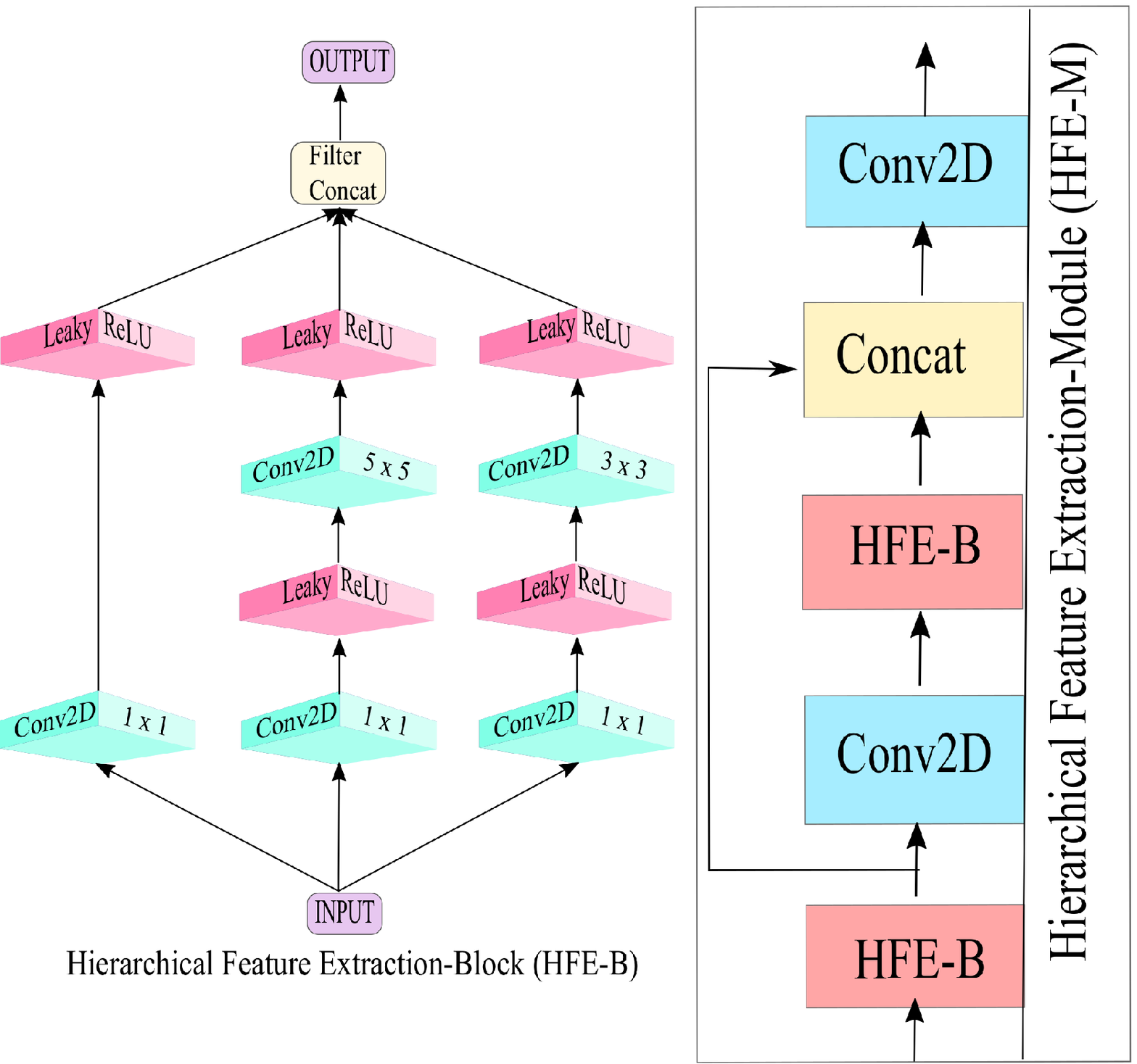}

\caption{Hierarchical Feature Extraction Module\label{fig:Hierarchical-Feature-Extraction}}
\end{figure}

Hierarchical Feature Extraction Module (HFE-M) is shown in figure \ref{fig:Hierarchical-Feature-Extraction}.
Hierarchical Feature Extraction Block (HFE-B) is the building block
of HFE-M. To sustain the long term memory dependency, residual
connections are used between the HFE-Ms. Total five HFE-Bs are used
in each HEF-CA phase, where the output feature maps of first HFE-B
are applied to the second block and so on. 

The motivation of Hierarchical Feature Extraction Module (HFE-M) is
taken from the inception architecture \cite{szegedy2015going}. As
depicted in the figure \ref{fig:Hierarchical-Feature-Extraction}a,
the same input is applied across three different convolution layers
with different kernel sizes. Notion for utilizing this complex filter
structure across the input is to acquire local as well as global features
of an input face image. Salient attributes like nose, eyes, lips,
ears and wrinkles of face images have distinct sizes across an image.
Therefore, to extract local attributes from an image, HFE-M uses convolution
layers with small kernel sizes ($1\times1$, $3\times3$). While the
hierarchical and the global features are extracted using larger kernel
sizes ($5\times5$). Before the convolution layer with large kernel
sizes ($3\times3$, $5\times5$), convolution layer with $1\times1$
kernel size is used to curb the input channels and hence reducing
the computational parameters of the architecture. All the convolution
layers are followed by LeakyReLU activation function in order to introduce
non-linearity in the model. The final feature maps are obtained by
concatenating the individual feature maps obtained from each convolution
layer with different kernel size. 

\subsubsection{Computationally Efficient Channel based Attention module}

\begin{figure}
\centering

\includegraphics[scale=0.35]{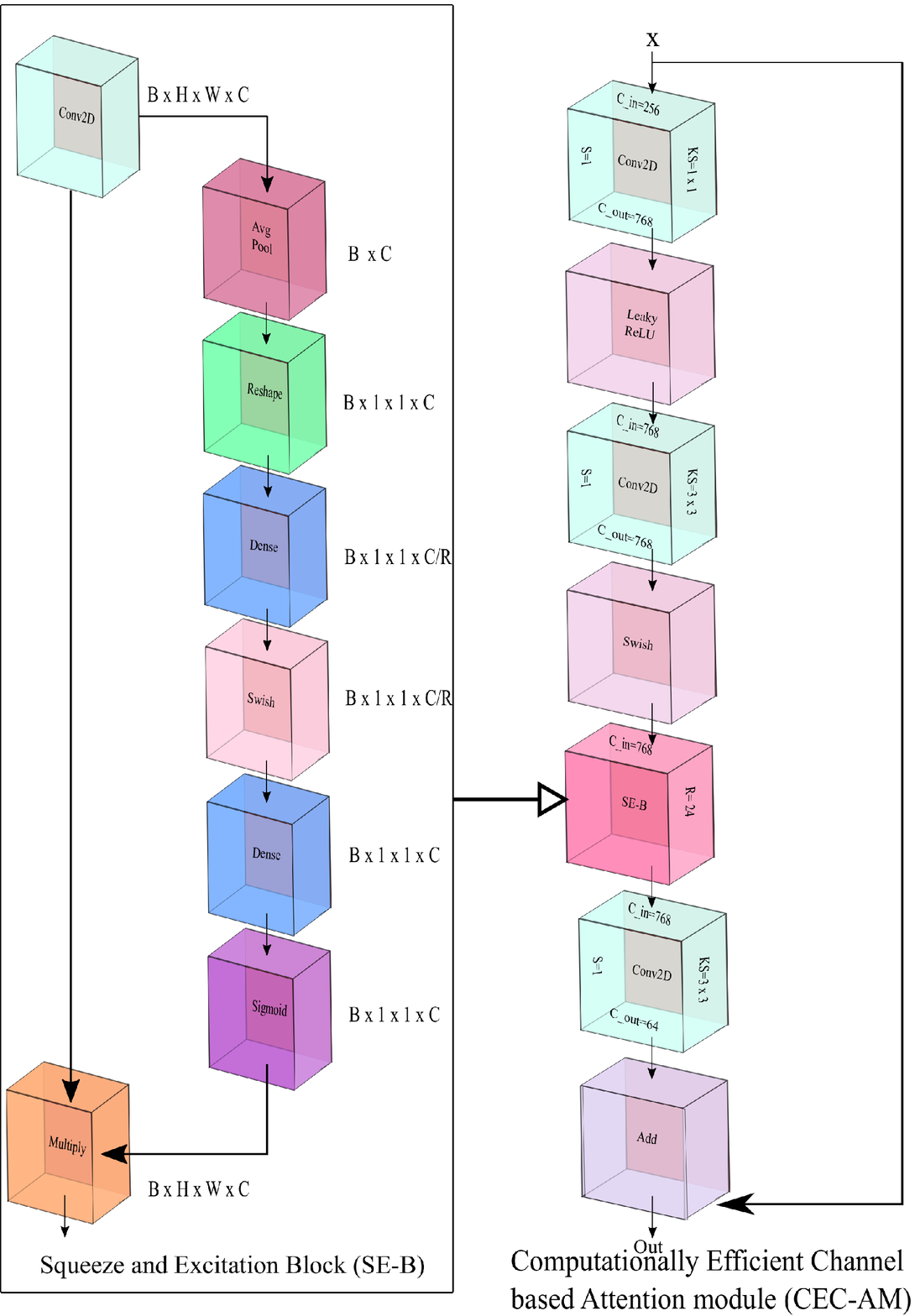}

\caption{Computationally Efficient Channel based Attention module\label{fig:Computationally-Efficient-Channe}}
\end{figure}

As depicted in figure \ref{fig:Computationally-Efficient-Channe},
the motivation for Computationally Efficient Channel based Attention Module (CEC-AM) is taken from mobilenet V3 model \cite{howard2019searching}.
Rather than using regular convolution layer, depthwise separable convolution
layers (combination of depthwise convolution layer and pointwise convolution
layer) are used in this block. In depthwise convolution, for each
channel in the feature space a single filter is applied and followed by $1\times1$
convolution using pointwise convolution layer to amalgamate
the feature maps of depthwise convolution layer. This layer is preferred
over the regular convolution layer due to its ability to use lesser
number of computational parameters without affecting the functionality
of traditional convolution layer. Second fundamental component used
in this module is squeeze and excitation block \cite{hu2018squeeze}.
The essence for using this block is to explicitly model the mutuality
present between the channels of convolution feature maps, guiding
the network to enhance the representational quality of output features.
Thus, CEC-AM is used to incline the architecture's ability to
assign the accessible processing resources to the most essential information
present in the input feature maps. 

\subsection{DCT based Auto Encoder Branch (DCTAE-B)}

As shown in figure \ref{fig:Generator-architecture} (sub-network1),
a Discrete Cosine Transform based Auto Encoder is employed in parallel
with PFH-B to incorporate frequency details in our super resolution
network. Basically, following two contributions are proposed in this
sub-network: 1) an auto encoder is employed in parallel with PFH-B to
generate high resolution DCT coefficients from low resolution DCT
coefficients. 2) DCT and IDCT blocks are defined in the network to
transform data from spatial domain to frequency domain and vice versa. 

The autoencoder takes the DCT coefficients of LR image as input and upsample
it to the DCT coefficients of HR scale. The use of skip-connections
ensure long passage of information in the network. We do not use any
form of normalization in the network as it tends to produce artifacts
in the image. The DCT to IDCT block helps in transforming the DCT
coefficients back to spatial domain, and thus provides a common link
between frequency and spatial domains. The output of AE is merged
with the output of PFH-B. This serves the purpose of using DCT coefficients
which have high frequency explicitly embedded in it.

\subsubsection{DCT and IDCT module}

\begin{figure}
\centering

\includegraphics[scale=0.6]{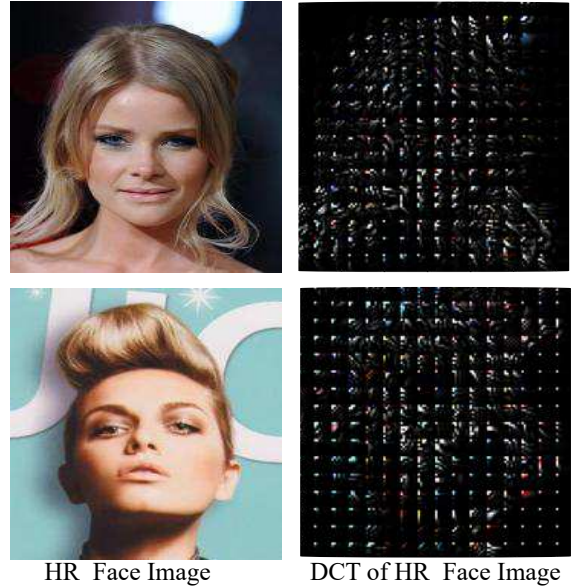}

\caption{Figure shows (from left to right) ground truth face image and corresponding Discrete Cosine Transform\label{fig:HR-face-image}}
\end{figure}

To operate in the frequency domain, firstly the face images are converted
from the spatial domain to frequency domain using DCT. For a single
block, DCT is calculated by the formula given in equation \ref{eq:dct}
and DCT of ground-truth image is shown in figure \ref{fig:HR-face-image}

\begin{align*}
D_{a,b}=\frac{1}{\sqrt{2M}}\beta\left(a\right)\beta\left(b\right)\stackrel[x=0]{M}{\sum}\stackrel[y=0]{M}{\sum}i_{x,y}\cos\left[\frac{\left(2x+1\right)a\pi}{2M}\right] \\
\cos\left[\frac{\left(2y+1\right)b\pi}{2M}\right] \numberthis
\label{eq:dct}
\end{align*}

here, block size is represented by $M$, image is denoted by $i$
and pixel coordinates as $x$ and $y.$ $a$ and $b$ represents indexes
of spatial frequency. Scale factor $\beta$ (refer eq. \ref{eq:dct_s})is used for transform to be orthogonal. 

\begin{math}
  {\beta\left(v\right)}=\left\{
    \begin{array}{ll}
      \nicefrac{1}{\sqrt{2}}, & \mbox{if $v=1$}.\\
      0, & \mbox{otherwise}.
    \end{array}
  \right.\label{eq:dct_s}
\end{math}

In order to establish a connection between the DCT based encoder network
and main SR network, IDCT is used. IDCT transforms the frequency domain
coefficients back to the spatial domain. The transformed values are
then fused with the main SR network to get output images with high
frequency details. The formula to calculate IDCT for single block
is given by equation \ref{eq:idct}

\begin{align*}
i_{x,y}=\frac{1}{\sqrt{2M}}\stackrel[x=0]{M}{\sum}\stackrel[y=0]{M}{\sum}\beta\left(a\right)\beta\left(b\right)D_{a,b}\cos\left[\frac{\left(2x+1\right)a\pi}{2M}\right] \\
\cos\left[\frac{\left(2y+1\right)b\pi}{2M}\right] \numberthis
\label{eq:idct}
\end{align*}

here, $i_{x,y}$ represents spatial domain image coefficients.

\subsection{Semantic Structural Constraint Branch}

As shown in figure \ref{fig:Generator-architecture} (subnetwork-2)
is SSC-B. This auxiliary branch guides the
our PFH-B to generate images with three dimensional parametric
feature information. This branch consists of consecutive five HFE-Bs
followed by a reconstruction module to achieve image size same as
of the final output image. 

To add structural semantic constraint in the proposed face super resolution,
we use 3D facial parameters fitted on a 2D image. In the following
subsection, we have explained the procedure to obtain the target
2D image fitted with 3D parametric information.

\subsubsection{3D model fitting to 2D images}

To obtain this image following steps are followed:
\begin{enumerate}
\item Firstly, facial landmarks of 2D facial image are
extracted through dlib library \cite{king2009dlib}.
\item We used PCA shape model of Surrey Face 3D Morphable model
to obtain a 3D face mesh constituting global and local face
parameters . Novel faces using PCA shape model are generated
by using formula mentioned in equation \ref{eq:1-1}
\begin{equation}
S=\bar{m}+\stackrel[i]{K}{\sum}\gamma_{i}\sigma_{i}m_{i}\label{eq:1-1}
\end{equation}
 here, $\bar{m}$ - mean of example meshes,$M=\left[m_{1},m_{2},....m_{n-1}\right]$
- set of principal components and $K$ denotes the total number of scans
utilized to make a model, $\gamma$ - PCA shape coefficients and $\sigma$
- Standard deviation
\item Using EOS library, the face mesh obtained above 
is fitted to the extracted landmarks \cite{huber2016multiresolution}. Four steps are followed to perform
this shape-to-landmark fitting: Estimate the pose of the facial image,
shape-specific identity fitting, linear expression fitting, and contour
(which includes front facial contour and occluding contour) fitting.
\item The last step is to render the obtained 3D face mesh parameters as
2D points on the face image as shown in figure \ref{fig:figure_3dmm}
\begin{figure}
\centering

\includegraphics[scale=0.3]{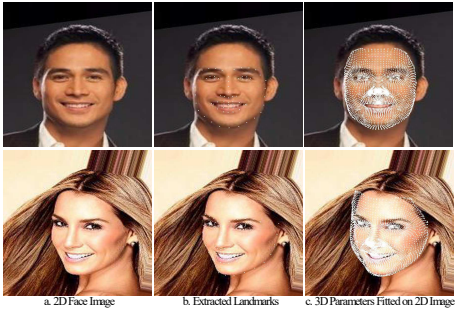}

\caption{\label{fig:figure_3dmm}Generation of training data for SSC-B. Figure shows (from left to right) a ground-truth 2D image, landmarks extraction on 2D images  and then 3D parameters fitting on 2D image.}
\end{figure}
\end{enumerate}

\subsection{Loss Functions}

To obtain the final output face image, loss function $L^{fh}$ is
optimized over $M$ training samples. Loss Function $L^{fh}$ is weighted
combination of loss functions explained in this section.

\subsubsection{DCT based loss function}

For subnetwork-1 i.e. autoencoder, $L_{1}$ loss between the generated
coefficients from subnetwork-1 and HR DCT coefficients of high resolution
face image is calculated. This loss function as this loss function
is trying to penalizes the proposed network for not predicting the high frequency
details correctly. The proposed DCT based loss function ($L_{dct/i.j}^{fh}$) is
defined in eq. \eqref{eq:dct_loss}

\begin{align*}
L_{dct/i.j}^{fh}=\frac{1}{W_{i,j}H_{i,j}}\stackrel[a=1]{W_{i,j}}{\sum}\stackrel[b=1]{H_{i,j}}{\sum}\mid\varDelta_{i,j}(i^{hr})_{a,b}- \\
(SN_{\Theta_{g1}}^{1}(\varDelta_{i,j}(i^{lr})))_{a,b} \numberthis \label{eq:dct_loss}
\end{align*}

where, $W$ and $H$ represents the dimensions of DCT based HR coefficients.
$\varDelta_{i,j}(i^{hr})$ and $\varDelta_{i,j}(i^{lr})$ are the
DCT coefficients extracted from the ground truth HR face image and
low resolution face image, respectively. $SN_{\Theta_{g}}^{1}$represents
the subnetwork-1 and its parameters.

\subsubsection{Semantic Structural Constraint Loss}

To add three dimensional parametric information to obtain the final
output image, we proposes semantic structural constraint loss. $L_{1}$
loss is calculated between the generated image from the subnetwork-2
and its corresponding ground-truth image (2D HR face image fitted
with 3D parametric information) and is given by eq. \ref{eq:3d_loss}

\begin{align*}
L_{ssc/i.j}^{fh}=\frac{1}{W_{i,j}H_{i,j}}\stackrel[a=1]{W_{i,j}}{\sum}\stackrel[b=1]{H_{i,j}}{\sum}\mid\varsigma_{i,j}(i^{hr})_{a,b}- r\\
(SN_{\Theta_{g2}}^{2}(i^{lr}))_{a,b}  \numberthis
\label{eq:3d_loss}
\end{align*}

here, $\varsigma_{i,j}(i^{hr})$ represents the 2D HR face image fitted
with 3D parametric information. $i^{lr}$ is the low resolution face
image which is passed to subnetwork-2 ($SN_{\Theta_{g2}}^{2}$) to
update its parameters.

\subsubsection{Final loss function}

Final loss function is the combination of dct based loss function,
semantic structural constraint loss, feature based $L_{2}$ loss and
adversarial loss. Feature based loss $L_{vgg}^{fh}$ and adversarial
losses $L_{adv}^{fh}$ are explained it detail by Ledig et al. \cite{ledig2017photo}.
So, final loss function ($L^{fh}$) is the combination of four loss
functions as mentioned in eq. \ref{eq:f_loss}
\begin{center}
\begin{equation}
L^{fh}=L_{vgg}^{fh}+\alpha L_{adv}^{fh}+\beta L_{dct}^{fh}+\gamma L_{ssc}^{fh}\label{eq:f_loss}
\end{equation}
\par\end{center}

here, $\alpha$, $\beta$ and $\gamma$ are the weight parameters
used to balance the impact of individual loss functions.

\section{Experiments}

\subsection{Datasets}

From CelebA dataset \cite{liu2015deep}, $108,640$ images are selected
for the training purpose, $5000$ images for validation and $5000$
for testing purpose. To validate the performance of proposed architecture,
we have performed experiments on benchmark face hallucination datasets-
Menpo dataset (left, right and semi-frontal profile) \cite{zafeiriou2017menpo}
and Helen dataset \cite{le2012interactive}.

\subsection{Implementation details}

The performance evaluation for the proposed architecture is performed for upscaling factors of $\times4$ and $\times8$ between the high
resolution face image and low resolution face image. For an upscaling
factor of $\times4$, the high resolution images are of size $128\times128$.
These images are downsampled using bicubic kernel with a factor of $\times4$
to generate a low resolution face images with size $32\times32$. For
an upscaling factor of $\times8$, the high resolution face images
($128\times128$) are downsampled with a factor of $\times8$ to generate
low resolution images ($16\times16$). 

For every stage in progressive face hallucination branch, five HFE-Ms
are used with fixed channel size ($64$) followed by a CEC-AM. In
CEC-AM, the number of input channels are $256$ and number of output
channels are 64 with an expansion factor of $3$. To reduce the number
of parameters in fully connected layer, total channels present is
a layer are divided by a factor $R$ having value $24$. The LeakyReLU hyper
parameter $\alpha$ is $0.2$. Batch size is
$8$; optimizer used is Adam with parameters $\beta_{1}=0.9$ and
$\beta_{2}=0.999$. Initial learning rate is set to $0.0001$. To
minimize the distance between the high resolution face image and generated
face image, $l^{fh}$ \ref{eq:f_loss} is used. To fully train the
model, alternate training between the generator and discriminator
is done to update their weights. Proposed model is evaluated using
two commonly used metrics: SSIM (Structural Similarity Index Measurement)
and PSNR (Peak Signal to Noise Ratio). For equitable comparison with
the previous works, these metrics are computed on Y-channel. 

\subsection{Ablation study}

In order to understand the importance of individual sub-modules of the proposed architecture, ablation study is conducted as summarized
in \ref{tab:Contribution-of-different}. 

\begin{figure*}[ht]
\centering

\includegraphics[angle=-90,scale=0.4]{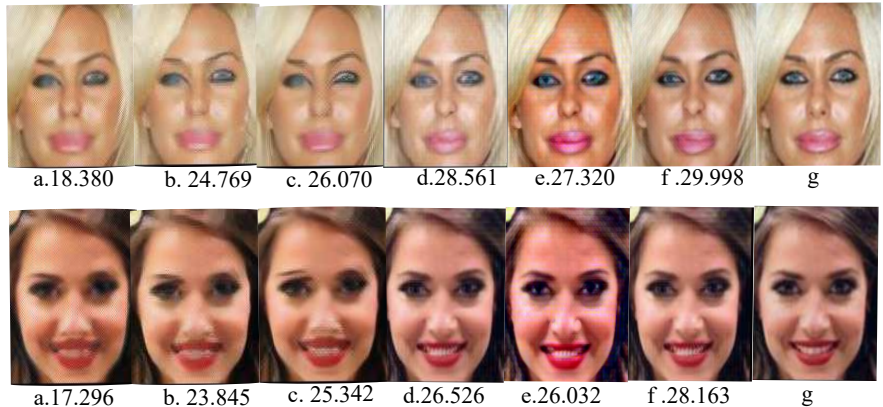}

\caption{\label{fig:Investigation-of-different}Investigation of different
components utilized in the proposed architecture: a) Single stage upscaling with HFE-M in generator, b) Multiscale upscaling with HFE-M, c) Adding Channel attention mechanism in b. d) Using sub-network1 (DCT based autoencoder) along with c. e) Using sub-network2 (Semantic Structural constraint block) along with c. f) Using both sub-network1 and sub-network2 along with c.}

\end{figure*}

Firstly, we studied the effect of using upsampling layer at different
positions in the architecture i.e. a single stage upsampling model
vs a progressive stage upsampling model. From the results obtained
after employing upsampling layer at end of the architecture (refer
figure \ref{fig:Investigation-of-different}a) and at progressive
stages (refer figure \ref{fig:Investigation-of-different}b), it is
clear that multi-stage upsampling performs better than single stage
upsampling model. As progressive upscaling approach allows the network
to mimic the fine details present in an input image and increases
its ability to learn. 

The proposed architecture is compared with or without using CEC-AM.
There is significant improvement in the results after adding this
module in the architecture (refer figure \ref{fig:Investigation-of-different}b
and \ref{fig:Investigation-of-different}c). This module is basically
used to slant the network's ability to provide access of available
resources to the most important information present in the feature
maps.

In order to further improve the quality of the generated image, we
embed sub-network1 in the architecture (refer \ref{tab:Contribution-of-different}d).
Results obtained (refer figure \ref{fig:Investigation-of-different}d)
after adding sub-network1 in the proposed architecture supports our
claim that DCT based auto encoder is able to add high frequency information
in the architecture.

Still some facial details like skin irregularities and depth information
is missing in the generated image. So, we tried to add these facial
details using sub-network2 (refer Table\ref{tab:Contribution-of-different}e).
In this experiment we used the progressive upscaling and sub-network2
and excluded the sub-network1. Results obtained shows substantial
improvement in the generated images perceptually. As facial details
and skin irregularities are more prominent in these images. But there
is little mismatch in the color as compare to the ground-truth images
(refer figure \ref{fig:Investigation-of-different}e and \ref{fig:Investigation-of-different}g).
So, for the final architecture we combined both the approaches i.e.
DCT based auto encoder and 2D images with 3D parametric information
to get the final output images (refer figure \ref{fig:Investigation-of-different}f),
performs better quantitatively and qualitatively. 

\begin{table}
\centering
\caption{Contribution of different components utilized in the proposed architecture\label{tab:Contribution-of-different}}
\begin{tabular}{|c|c|c|c|c|c|c|}
\hline 
{\small{}Components} & {\small{}a} & {\small{}b} & {\small{}c} & {\small{}d} & {\small{}e} & {\small{}f}\tabularnewline
\hline 
\hline 
{\small{}Single stage} & {\small{}$\checked$} &  &  &  &  & \tabularnewline
\hline 
{\small{}Multiple stages} &  & {\small{}$\checked$} & {\small{}$\checked$} & {\small{}$\checked$} & {\small{}$\checked$} & {\small{}$\checked$}\tabularnewline
\hline 
{\small{}HFE-M} & {\small{}$\checked$} & {\small{}$\checked$} & {\small{}$\checked$} & {\small{}$\checked$} & {\small{}$\checked$} & {\small{}$\checked$}\tabularnewline
\hline 
{\small{}CEC-AM} &  &  & {\small{}$\checked$} & {\small{}$\checked$} & {\small{}$\checked$} & {\small{}$\checked$}\tabularnewline
\hline 
{\small{}Sub-network1} &  &  &  & {\small{}$\checked$} &  & {\small{}$\checked$}\tabularnewline
\hline 
{\small{}Sub-network2} &  &  &  &  & {\small{}$\checked$} & {\small{}$\checked$}\tabularnewline
\hline 
\end{tabular}

\end{table}

\subsection{Comparison with the state-of-the-art method}

We compared our proposed architecture with seven state-of-the-art
methods: SRCNN \cite{dong2014learning}, VDSR \cite{kim2016accurate},
SRGAN \cite{ledig2017photo}, ESRGAN \cite{wang2018esrgan}, ImprovedFSR
\cite{wang2020improved}, SICNN \cite{zhang2018super}, SAM3D \cite{hu2020face}
and bicubic interpolation to show the efficacy of our network. For
fair comparison, we trained all these models on our training dataset
with same parameters. 

\textbf{Menpo dataset}- We evaluated the performance of our model
and other state-of-the-art methods on Menpo dataset (left, right and
semi-frontal profiles) \cite{zafeiriou2017menpo} qualitatively and quantitatively. For left
profile, our model has achieved second highest PSNR and highest SSIM
for both $\times4$ and $\times8$ scaling factors (refer Table \ref{tab:Quantitative-result-comparison-1}).
For right profile our model has achieved highest PSNR and SSIM for
$\times4$ scale. For $\times8$ scale, our model has achieved second
highest PSNR and highest SSIM values. For semi-frontal profile $\times4$
scale, our model has second highest PSNR and highest SSIM values and
highest PSNR and SSIM numbers for $\times8$ scale. 

\begin{figure}
\centering

\includegraphics[scale=0.3]{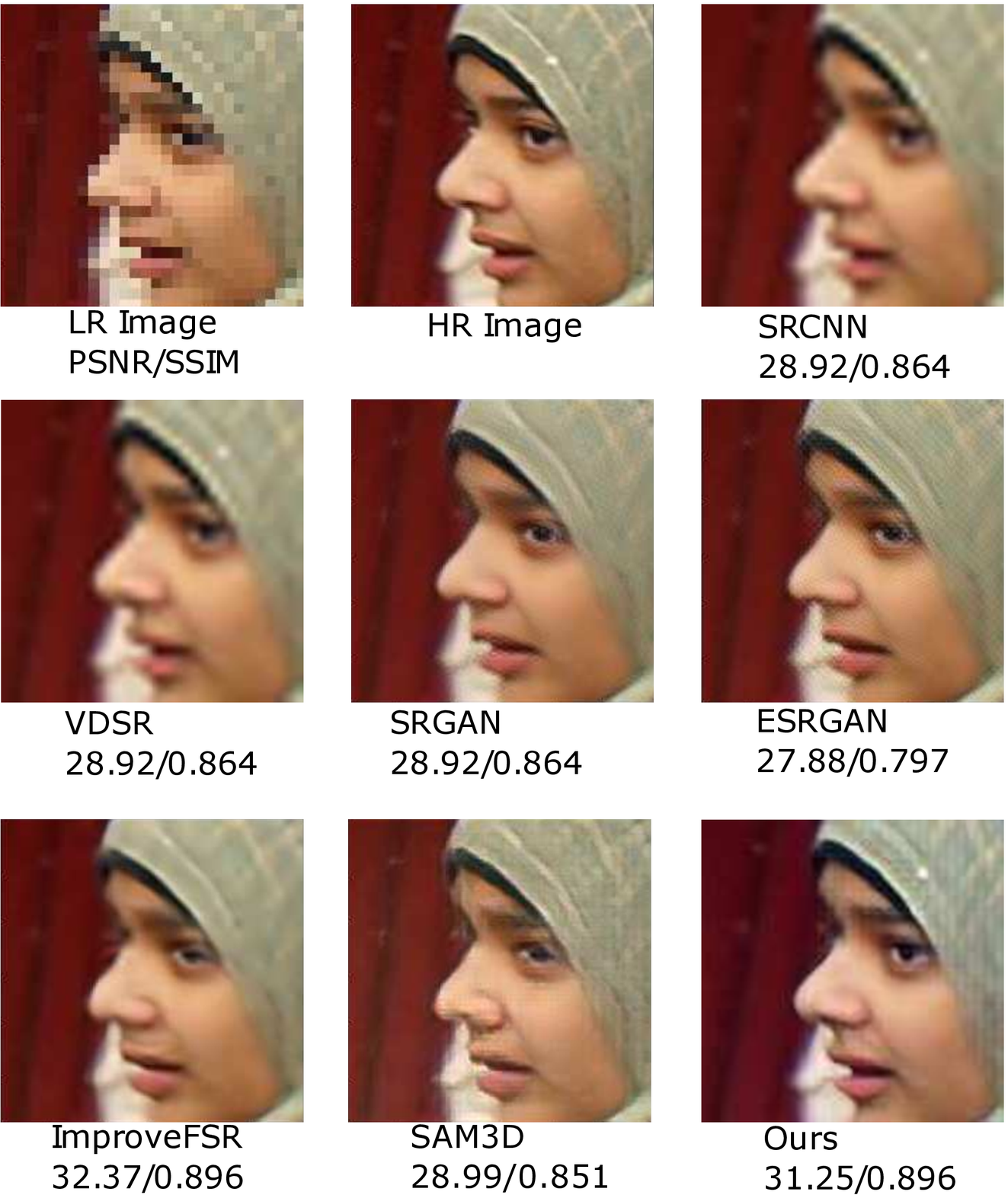}\caption{Perceptual and quantitative (PSNR/SSIM) result comparison with state-of-the-art
methods for the magnification factor of $\times4$ on Menpo test dataset.\label{fig:Perceptual-and-quantitative}}
\end{figure}

\begin{figure}
\centering

\includegraphics[scale=0.3]{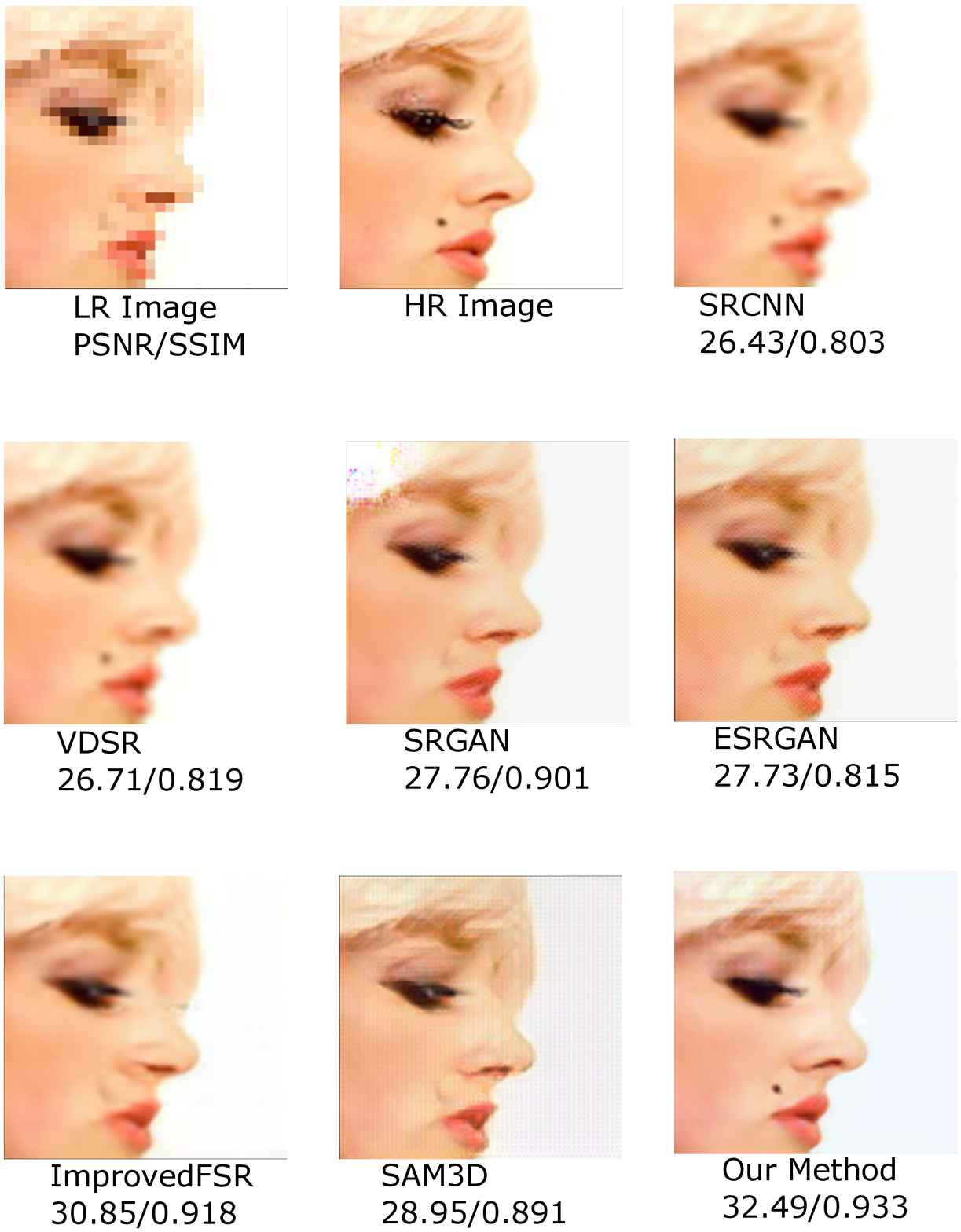}

\caption{Perceptual and quantitative (PSNR/SSIM) result comparison with state-of-the-art
methods for the magnification factor of $\times4$ on Menpo test dataset.\label{fig:Perceptual-and-quantitative-1}}

\end{figure}

\begin{figure}
\centering

\includegraphics[scale=0.3]{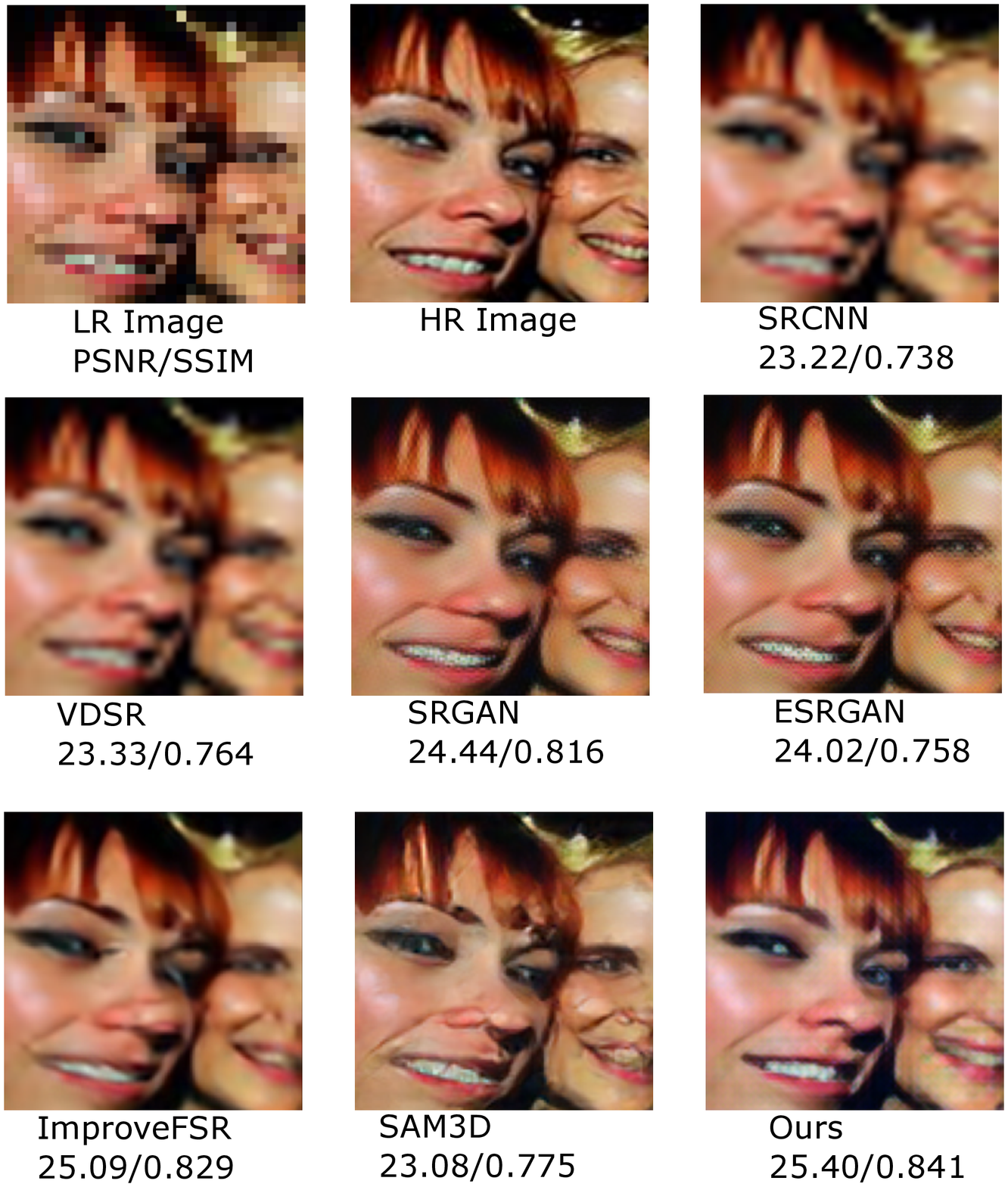}

\caption{Perceptual and quantitative (PSNR/SSIM) result comparison with state-of-the-art
methods for the magnification factor of $\times4$ on Menpo test dataset.\label{fig:Perceptual-and-quantitative-2}}

\end{figure}

Perceptual results for Menpo dataset are represented in figures \ref{fig:Perceptual-and-quantitative},
\ref{fig:Perceptual-and-quantitative-1} and \ref{fig:Perceptual-and-quantitative-2}.
In figure \ref{fig:Perceptual-and-quantitative}, although the image
generated by ImprovedFSR has highest PSNR, still there are artifacts
present in the eyes and mouth region. In SAM3D model, the generated
image has blemished skin with noise present in the mouth region. Only
the image generated by proposed model is able to mimic the ground-truth
image. As shown in figure \ref{fig:Perceptual-and-quantitative-1},
the image generated by SRGAN model has some artifacts in the hair
region. And none of the competing GAN methods are able to generate
a mole present on the face. Our method is recovering small details
or textures present in the face image. In figure \ref{fig:Perceptual-and-quantitative-2},
competing GAN methods are unable to recover the details of the nose
region. And CNN based methods are producing smooth faces with very
less textural details. Image generated by our model is looking perceptually
better than all other methods.

\begin{table*}[ht]
\centering
\caption{Quantitative result comparison on the basis of average PSNR (dB) and
average SSIM on different facial poses (left, right and semi-frontal)
of Menpo dataset.\label{tab:Quantitative-result-comparison-1}}
{\scriptsize{}}%
\begin{tabular}{|c|c|c|c|c|c|c|c|c|c|c|c|c|}
\hline 
{\scriptsize{}Scale} & \multicolumn{6}{c|}{{\scriptsize{}$\times4$}} & \multicolumn{6}{c}{{\scriptsize{}$\times8$}}\tabularnewline
\hline 
 & \multicolumn{2}{c|}{{\scriptsize{}Left}} & \multicolumn{2}{c|}{{\scriptsize{}Right}} & \multicolumn{2}{c|}{{\scriptsize{}Semi-frontal}} & \multicolumn{2}{c|}{{\scriptsize{}Left}} & \multicolumn{2}{c|}{{\scriptsize{}Right}} & \multicolumn{2}{c|}{{\scriptsize{}Semi-frontal}}\tabularnewline
\hline 
\hline 
 & {\scriptsize{}PSNR} & {\scriptsize{}SSIM} & {\scriptsize{}PSNR} & {\scriptsize{}SSIM} & {\scriptsize{}PSNR} & {\scriptsize{}SSIM} & {\scriptsize{}PSNR} & {\scriptsize{}SSIM} & {\scriptsize{}PSNR} & {\scriptsize{}SSIM} & {\scriptsize{}PSNR} & {\scriptsize{}SSIM}\tabularnewline
\hline 
{\scriptsize{}Bicubic} & {\scriptsize{}27.32} & {\scriptsize{}0.794} & {\scriptsize{}26.28} & {\scriptsize{}0.772} & {\scriptsize{}24.89} & {\scriptsize{}0.763} & {\scriptsize{}22.01} & {\scriptsize{}0.634} & {\scriptsize{}21.07} & {\scriptsize{}0.602} & {\scriptsize{}20.31} & {\scriptsize{}0.501}\tabularnewline
\hline 
{\scriptsize{}SRCNN\cite{dong2014learning}} & {\scriptsize{}26.99} & {\scriptsize{}0.811} & {\scriptsize{}26.59} & {\scriptsize{}0.794} & {\scriptsize{}25.15} & {\scriptsize{}0.772} & {\scriptsize{}22.14} & {\scriptsize{}0.663} & {\scriptsize{}21.16} & {\scriptsize{}0.623} & {\scriptsize{}20.99} & {\scriptsize{}0.509}\tabularnewline
\hline 
{\scriptsize{}VDSR\cite{kim2016accurate}} & {\scriptsize{}27.45} & {\scriptsize{}0.831} & {\scriptsize{}26.78} & {\scriptsize{}0.793} & {\scriptsize{}25.32} & {\scriptsize{}0.769} & {\scriptsize{}22.56} & {\scriptsize{}0.671} & {\scriptsize{}21.12} & {\scriptsize{}0.625} & {\scriptsize{}20.61} & {\scriptsize{}0.512}\tabularnewline
\hline 
{\scriptsize{}SRGAN\cite{ledig2017photo}} & {\scriptsize{}29.34} & {\scriptsize{}0.873} & {\scriptsize{}29.51} & {\scriptsize{}0.871} & {\scriptsize{}28.94} & {\scriptsize{}0.868} & {\scriptsize{}22.34} & {\scriptsize{}0.681} & {\scriptsize{}22.31} & {\scriptsize{}0.661} & {\scriptsize{}21.02} & {\scriptsize{}0.557}\tabularnewline
\hline 
{\scriptsize{}ESRGAN\cite{wang2018esrgan}} & {\scriptsize{}28.99} & {\scriptsize{}0.821} & {\scriptsize{}28.83} & {\scriptsize{}0.813} & {\scriptsize{}28.43} & {\scriptsize{}0.813} & {\scriptsize{}21.04} & {\scriptsize{}0.662} & {\scriptsize{}21.06} & {\scriptsize{}0.621} & {\scriptsize{}20.19} & {\scriptsize{}0.532}\tabularnewline
\hline 
{\scriptsize{}SICNN\cite{zhang2018super}} & {\scriptsize{}28.09} & {\scriptsize{}0.812} & {\scriptsize{}28.12} & {\scriptsize{}0.803} & {\scriptsize{}27.02} & {\scriptsize{}0.799} & {\scriptsize{}22.45} & {\scriptsize{}0.679} & {\scriptsize{}22.69} & {\scriptsize{}0.659} & {\scriptsize{}21.21} & {\scriptsize{}0.613}\tabularnewline
\hline 
{\scriptsize{}ImprovedFSR\cite{wang2020improved}} & \textbf{\scriptsize{}30.85} & {\scriptsize{}0.887} & {\scriptsize{}30.78} & {\scriptsize{}0.884} & \textbf{\scriptsize{}30.25} & {\scriptsize{}0.877} & \textbf{\scriptsize{}23.90} & {\scriptsize{}0.713} & \textbf{\scriptsize{}23.56} & {\scriptsize{}0.676} & {\scriptsize{}21.28} & {\scriptsize{}0.623}\tabularnewline
\hline 
{\scriptsize{}SAM3D\cite{hu2020face}} & {\scriptsize{}28.23} & {\scriptsize{}0.842} & {\scriptsize{}28.91} & {\scriptsize{}0.848} & {\scriptsize{}27.92} & {\scriptsize{}0.838} & {\scriptsize{}23.71} & {\scriptsize{}0.692} & {\scriptsize{}23.01} & {\scriptsize{}0.654} & {\scriptsize{}21.11} & {\scriptsize{}0.601}\tabularnewline
\hline 
{\scriptsize{}Ours} & {\scriptsize{}30.36} & \textbf{\scriptsize{}0.924} & \textbf{\scriptsize{}30.81} & \textbf{\scriptsize{}0.925} & {\scriptsize{}29.80} & \textbf{\scriptsize{}0.922} & {\scriptsize{}23.73} & \textbf{\scriptsize{}0.723} & {\scriptsize{}23.52} & \textbf{\scriptsize{}0.689} & \textbf{\scriptsize{}21.34} & \textbf{\scriptsize{}0.641}\tabularnewline
\hline 
\end{tabular}{\scriptsize\par}

\end{table*}

\begin{figure}
\centering

\includegraphics[scale=0.3]{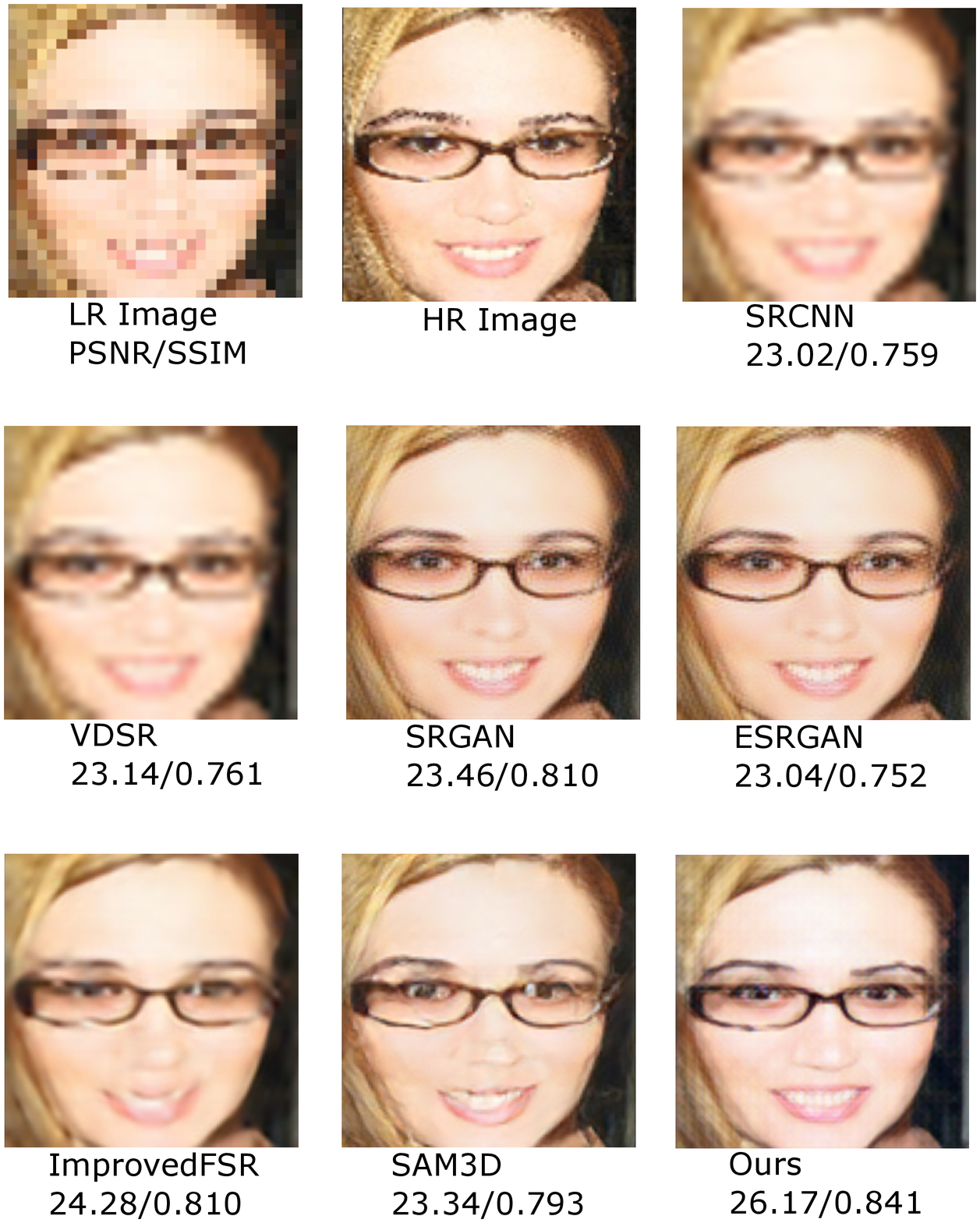}

\caption{Perceptual and quantitative (PSNR/SSIM) result comparison with state-of-the-art
methods for the magnification factor of $\times4$ on Helen test dataset.\label{fig:Perceptual-and-quantitativehelen}}
\end{figure}

\textbf{Helen dataset}- For Helen test dataset \cite{le2012interactive}, quantitative results
(PSNR/SSIM) are shown in Table \ref{tab:Quantitative-result-comparison}
for an upscaling factor of $\times4$. Our method is able to achieve
hightest PSNR and SSIM values as compare to state-of-the-art methods.
Perceptual analysis with the PSNR and SSIM values is presented in
figure \ref{fig:Perceptual-and-quantitativehelen}. CNN based methods
like SRCNN and VDSR are generating output images with blurriness.
SRGAN and ESRGAN are also producing images with some artifacts at
the eyes and cheeks. Other competing methods like ImprovedFSR and
SAM3D are unable to generate better images perceptually. Only the proposed
method is able to recover fine textural details like eye brows and
depth details similar to the ground-truth image. 

\begin{figure}
\centering

\includegraphics[scale=0.3]{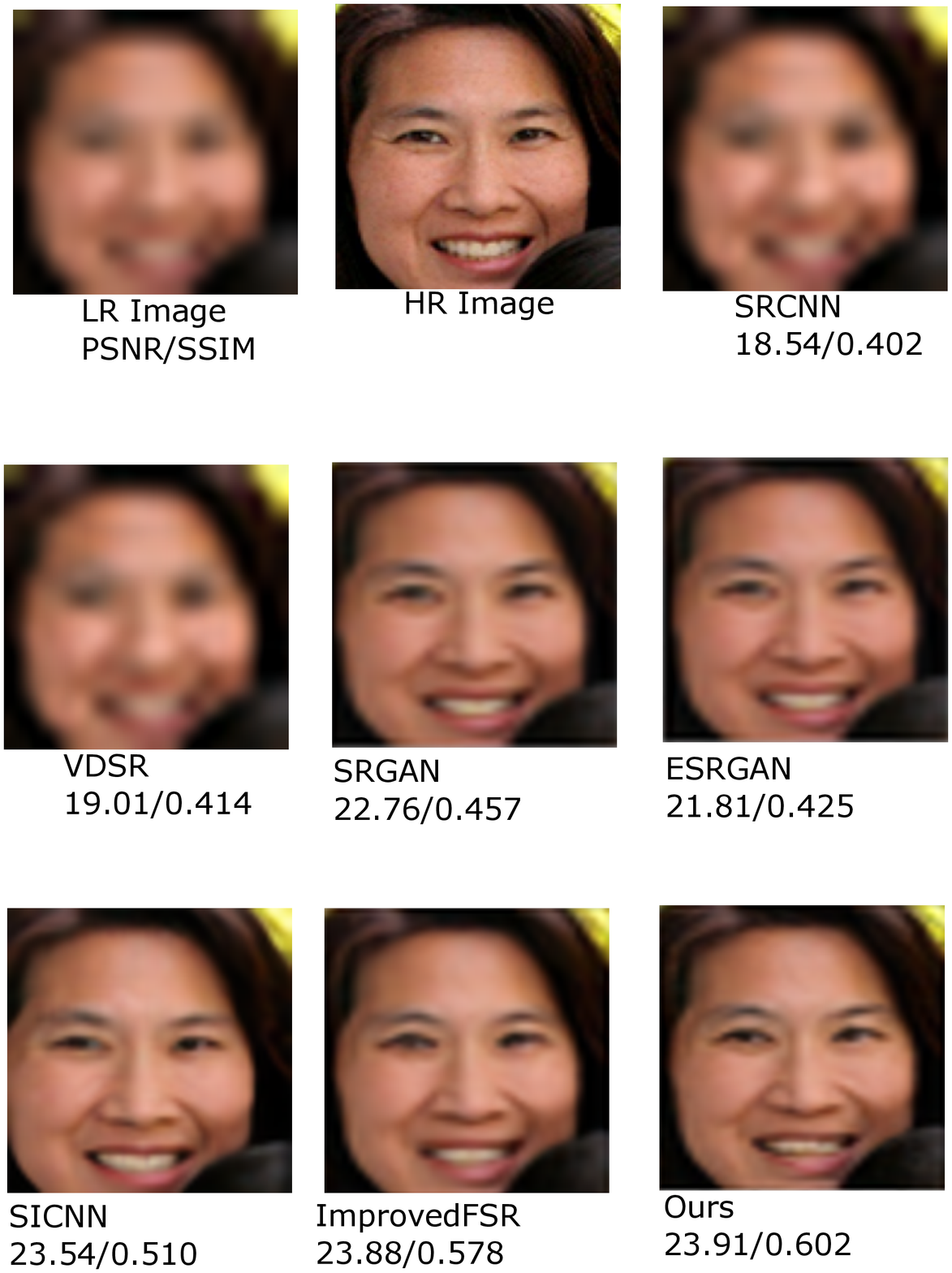}

\caption{Perceptual and quantitative (PSNR/SSIM) result comparison with state-of-the-art
methods for the magnification factor of $\times8$ on Helen test dataset.\label{fig:Perceptual-and-quantitativehelen1}}

\end{figure}

\begin{figure}
\centering

\includegraphics[scale=0.3]{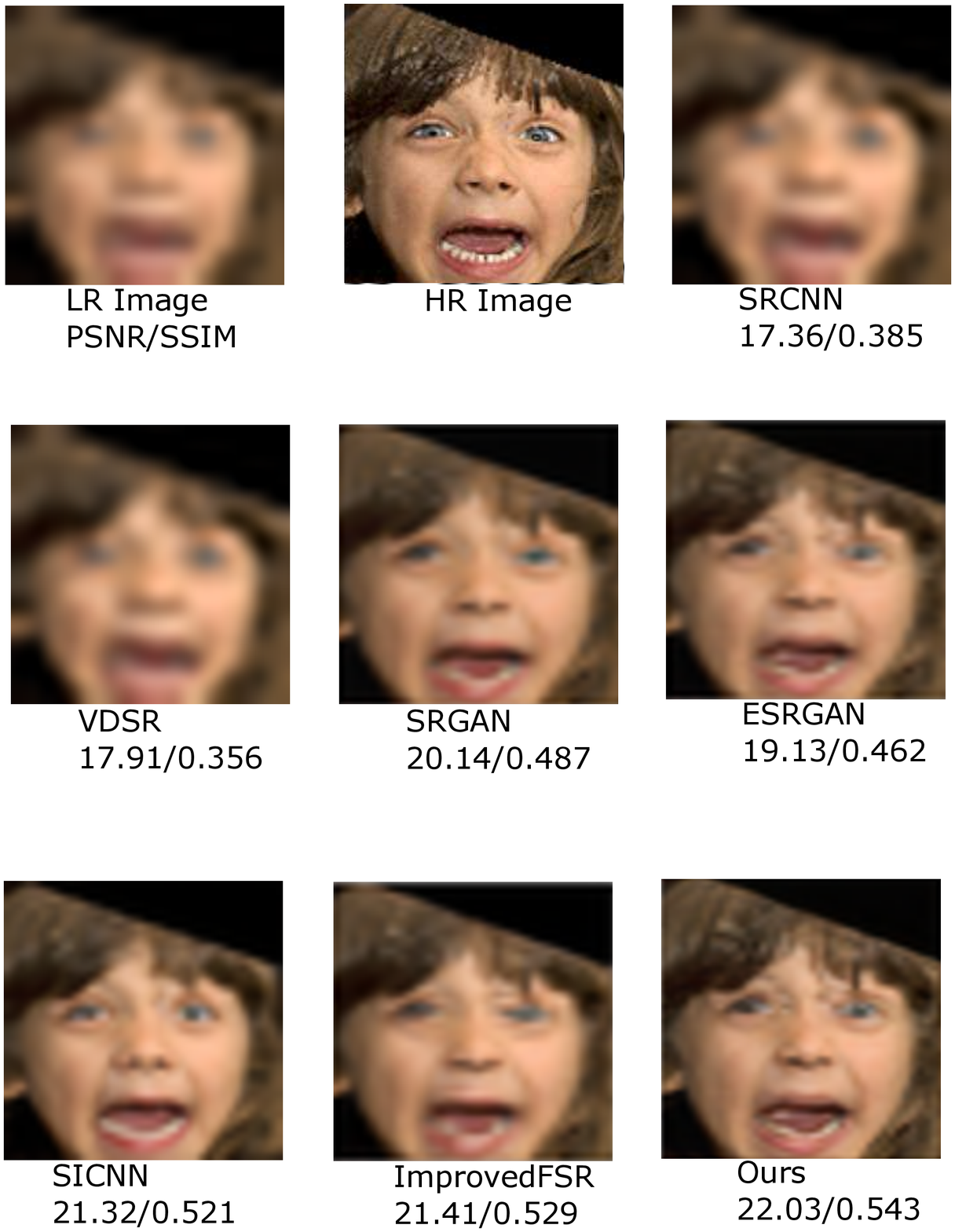}

\caption{Perceptual and quantitative (PSNR/SSIM) result comparison with state-of-the-art
methods for the magnification factor of $\times8$ on Helen test dataset.\label{fig:Perceptual-and-quantitativehelen2}}

\end{figure}

Proposed model has achieved highest SSIM and second highest PSNR for $\times8$
factor (refer Table \ref{tab:Quantitative-result-comparison}). Qualitative
analysis with their quantitative numbers are presented in figures
\ref{fig:Perceptual-and-quantitativehelen1} and \ref{fig:Perceptual-and-quantitativehelen2}.
From the figures, it is clear that perceptually our model is performing
better than the other competing methods. ImprovedFSR has highest PSNR
value but visually our model is able to recover more finer details
and textures. 

\begin{table}
\centering
\caption{Quantitative result comparison on the basis of average PSNR (dB) and
average SSIM of Helen dataset.\label{tab:Quantitative-result-comparison}}
\begin{tabular}{|c|c|c|c|c|}
\hline 
\scriptsize{}Scale & \multicolumn{2}{c|}{\scriptsize{}$\times4$} & \multicolumn{2}{c|}{\scriptsize{}$\times8$}\tabularnewline
\hline 
\hline 
 & {\scriptsize{}PSNR} & {\scriptsize{}SSIM} & {\scriptsize{}PSNR} & {\scriptsize{}SSIM}\tabularnewline
\hline 
{\scriptsize{}Bicubic} & {\scriptsize{}25.06} & {\scriptsize{}0.692} & {\scriptsize{}21.67} & {\scriptsize{}0.612}\tabularnewline
\hline 
{\scriptsize{}SRCNN}\cite{dong2014learning} & {\scriptsize{}26.45} & {\scriptsize{}0.712} & {\scriptsize{}22.04} & {\scriptsize{}0.634}\tabularnewline
\hline 
{\scriptsize{}VDSR}\cite{kim2016accurate} & {\scriptsize{}26.89} & {\scriptsize{}0.734} & {\scriptsize{}22.14} & {\scriptsize{}0.639}\tabularnewline
\hline 
{\scriptsize{}SRGAN}\cite{ledig2017photo} & {\scriptsize{}28.45} & {\scriptsize{}0.851} & {\scriptsize{}22.31} & {\scriptsize{}0.689}\tabularnewline
\hline 
{\scriptsize{}ESRGAN}\cite{wang2018esrgan} & {\scriptsize{}27.86} & {\scriptsize{}0.790} & {\scriptsize{}21.99} & {\scriptsize{}0.674}\tabularnewline
\hline 
{\scriptsize{}SICNN}\cite{zhang2018super} & {\scriptsize{}26.43} & {\scriptsize{}0.757} & {\scriptsize{}22.76} & {\scriptsize{}0.681}\tabularnewline
\hline 
{\scriptsize{}ImprovedFSR}\cite{wang2020improved} & {\scriptsize{}28.83} & {\scriptsize{}0.856} & {\scriptsize{}\textbf{23.99}} & {\scriptsize{}0.701}\tabularnewline
\hline 
{\scriptsize{}SAM3D}\cite{hu2020face} & {\scriptsize{}27.32} & {\scriptsize{}0.834} & {\scriptsize{}22.16} & {\scriptsize{}0.712}\tabularnewline
\hline 
{\scriptsize{}Ours} & {\scriptsize{}\textbf{28.86}} & {\scriptsize{}\textbf{0.911}} & {\scriptsize{}23.83} & {\scriptsize{}\textbf{0.741}}\tabularnewline
\hline 
\end{tabular}
\end{table}

\section{Conclusion}

Current work presents a novel GAN based progressive face hallucination
network. To generate the final output image with 3D parametric information,
proposed model uses a auxiliary supervision network which is compelled
to generate 2D images with 3D parametric information using shape model
of 3DMM. To incorporate high frequency components in the image, an
auto encoder is proposed which generates high resolution coefficients
of DCT. To embed high resolution DCT information into the face hallucination
network IDCT block is introduced within the network to convert the
frequency domain coefficients to spatial domain. Output images generated
by the proposed model have subtle structural details with depth information,
outperforming the state-of-the-art methods.

\section*{Acknowledgments}

We are greateful to Harsh Vardhan Dogra for insighful discussions.

\ifCLASSOPTIONcaptionsoff
  \newpage
\fi



%
\bibliographystyle{IEEEtran}
\bibliography{REFERENCES}

\end{document}